\documentclass{article}

    \usepackage[preprint, nonatbib]{neurips_2023}

\usepackage[utf8]{inputenc} 
\usepackage[T1]{fontenc}    
\usepackage{hyperref}       
\usepackage{url}            
\usepackage{booktabs}       
\usepackage{amsfonts}       
\usepackage{nicefrac}       
\usepackage{microtype}      
\usepackage[table]{xcolor}         
\usepackage{amsmath}
\usepackage{multirow}
\usepackage[export]{adjustbox}
\usepackage{subcaption}
\usepackage{subfiles}
\usepackage{pifont}
\usepackage{enumitem}
\usepackage{pdfpages}

\def\ie{{\it i.e. }}
\def\eg{{\it e.g. }}
\def\vs{{\it vs. }}

\definecolor{myred}{RGB}{197, 90, 17}
\definecolor{mygreen}{RGB}{112, 173, 71}

\DeclareMathOperator*{\argmax}{arg\,max}

\newcommand{\xmark}{\ding{55}}%
\newcommand{\MyMapTemplatePrefixc}[4]{\expandafter#1\csname#3#4\endcsname{#2{#4}}} 
\forcsvlist{\MyMapTemplatePrefixc {\def} {\mathcal}{c}} {A,B,C,D,E,F,G,H,I,J,K,L,M,N,O,P,Q,R,S,T,U,V,W,X,Y,Z}  

\newcommand{\MyMapTemplateNoPrefix}[3]{\expandafter#1\csname#3\endcsname{#2{#3}}}
\forcsvlist{\MyMapTemplateNoPrefix {\def} {\mathbf} } {0,1,a,b,c,d,e, f, g, h, i, j, k, l, m, n, o, p, q, r, s, t, u, v, w, x, y, z} 
\forcsvlist{\MyMapTemplateNoPrefix {\def} {\mathbf} } {A,B,C,D,E,F,G,H,I,J,K,L,M,N,O,P,Q,R,S,T,U,V,W,X,Y,Z}  

\title{Prompting Diffusion Representations for Cross-Domain Semantic Segmentation}

\author{%
  Rui Gong\textsuperscript{\rm 1}   \space\space   Martin Danelljan\textsuperscript{\rm 1}     \space\space    Han Sun\textsuperscript{\rm 2}    \space\space  Julio Delgado Mangas\textsuperscript{\rm 3}    \space\space   Luc Van Gool\textsuperscript{\rm 1}\\
\textsuperscript{\rm 1}CVL, ETH Z\"urich \textsuperscript{\rm 2} IMOS, EPFL \textsuperscript{\rm 3} Meta Reality Labs\\
{\texttt{\{gongr,martin.danelljan,vangool\}@vision.ee.ethz.ch}}\\
{\texttt{han.sun@epfl.ch julio.delgadomangas@gmail.com}}\\
}

\begin{document}

\maketitle

\begin{abstract}
While originally designed for image generation, diffusion models have recently shown to provide excellent pretrained feature representations for semantic segmentation. 
Intrigued by this result, we set out to explore how well diffusion-pretrained representations generalize to new domains, a crucial ability for any representation.
We find that diffusion-pretraining achieves extraordinary domain generalization results for semantic segmentation, outperforming both supervised and self-supervised backbone networks.
Motivated by this, we investigate how to utilize the model's unique ability of taking an input prompt, in order to further enhance its cross-domain performance. We introduce a scene prompt and a prompt randomization strategy to help further disentangle the domain-invariant information when training the segmentation head. Moreover, we propose a simple but highly effective approach for test-time domain adaptation, based on learning a scene prompt on the target domain in an unsupervised manner. Extensive experiments conducted on four synthetic-to-real and clear-to-adverse weather benchmarks demonstrate the effectiveness of our approaches. 
Without resorting to any complex techniques, such as image translation, augmentation, or rare-class sampling, we set a new state-of-the-art on all benchmarks. Our implementation will be publicly available at \url{https://github.com/ETHRuiGong/PTDiffSeg}.

\end{abstract}

\section{Introduction}
Deep neural networks for semantic segmentation have achieved remarkable performance when trained and tested on the data from the same distribution~\cite{long2015fully,chen2017deeplab,wang2020deep,xie2021segformer}. However, their ability to generalize to new and diverse data remains limited~\cite{tsai2018learning,vu2018advent,zou2018unsupervised,tranheden2021dacs}. Deep semantic segmentation models are sensitive to domain shifts, which occurs when the distribution of the testing (target) data differs from that of the training (source) data. This often leads to drastic performance degradation. To enhance the generalization ability of deep models to unseen scenarios, domain generalization (DG) methods employ specialized training strategies that improve the model's robustness. Additionally, as an alternative to DG, test-time domain adaptation (TTDA) aims to adapt a model trained on the source domain by only utilizing unlabelled target domain data.

Diffusion models have recently achieved extraordinary results for image generation and synthesis tasks~\cite{ho2020denoising,rombach2022high,ramesh2022hierarchical,hertz2022prompt}.
At the heart of the diffusion model lies the idea of training a denoising autoencoder to learn the reverse of a Markovian diffusion process. Trained on large-scale paired image-text datasets like LAION5B~\cite{schuhmann2022laion}, diffusion models, such as Stable-Diffusion~\cite{rombach2022high}, have demonstrated remarkable performance on image synthesis controlled by natural language. 
The ability of large-scale text-to-image diffusion models to produce visually stunning images with intricate details, varied content, and coherent structures, while retaining the ability to modify and compose semantics, is a remarkable breakthrough. 
It implies that the diffusion models implicitly learn both high-level and low-level visual representations from the vast collections of image-text pairs dataset.
Recently, frozen diffusion models have therefore been shown to provide excellent feature representations for semantic segmentation~\cite{zhao2023unleashing,xu2023open}, providing an alternative to standard supervised ImageNet~\cite{deng2009imagenet} or self-supervised pretraining~\cite{he2022masked}.

\begin{figure*}
    \centering
    \includegraphics[width=\linewidth]{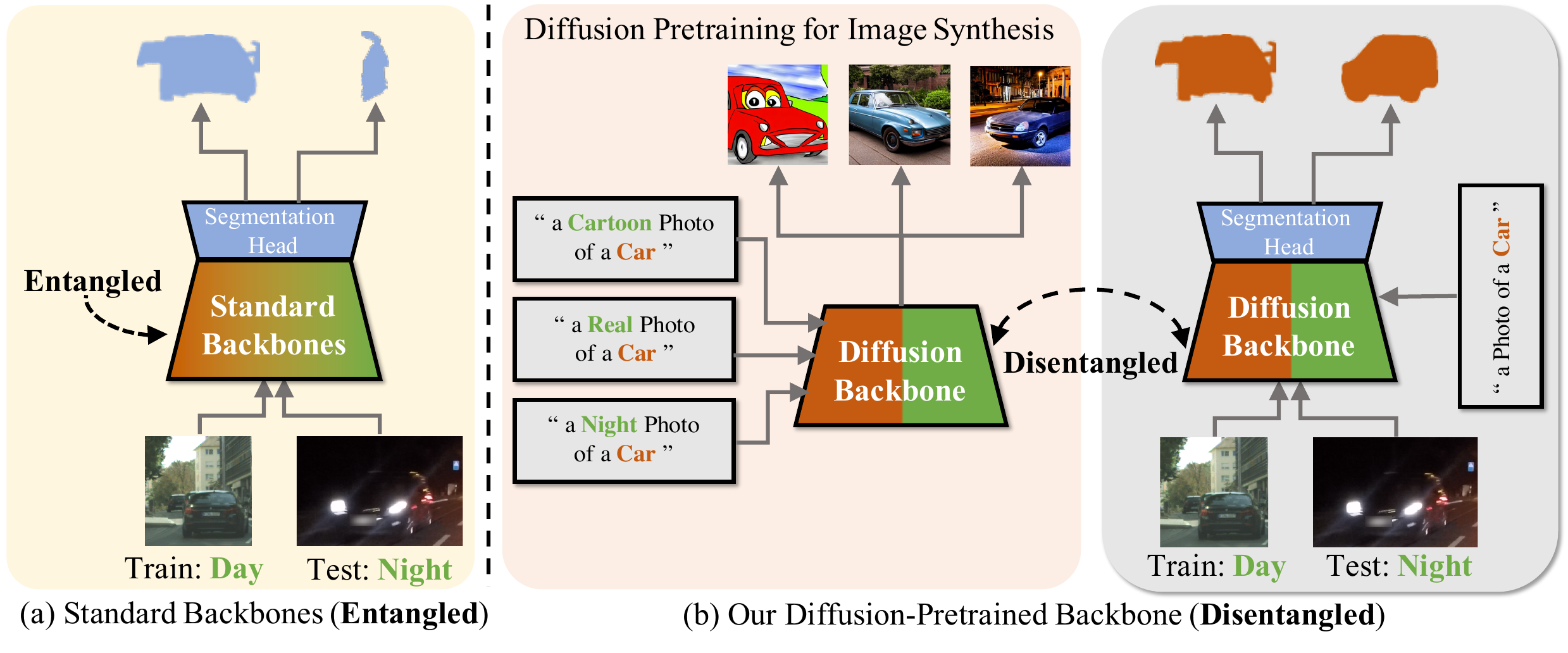}
    \vspace{-20pt}
    \caption{\textbf{Standard \textit{\textbf{vs.}} Diffusion Backbones.} (a) In standard backbones, the \textbf{\textcolor{myred}{domain-invariant}} (\eg semantics) and \textbf{\textcolor{mygreen}{domain-variant}} (\eg styles, lighting) information are partially entangled in the latent visual representations. Consequently, semantic segmentation models trained on these backbones often suffer from significant performance degradation when confronted with domain shifts. (b) Conversely, diffusion models naturally learn a more disentangled representation through text-to-image synthesis pretraining, as it is capable of generating the same semantic content under diverse styles.
    When used as a backbone network for semantic segmentation, it therefore offers a feature representation robust to large domain shifts, leading to superior generalization capabilities across diverse domains.}
    \label{fig:motivation}
    \vspace{-10pt}
\end{figure*}

In light of the success of diffusion models for supervised semantic segmentation, we are led to contemplate: \emph{How well do diffusion-pretrained semantic segmentation models generalize to unseen domains?} In this paper, we first investigate this question by comparing the generalization performance of diffusion-pretraining with other popular backbones and pretraining approaches. 
We find that the vanilla diffusion models show exceptional generalization ability, surpassing that of all other backbones.
We attribute this to the natural disentanglement of concepts that occur in diffusion models. As illustrated in Fig.~\ref{fig:motivation}(b), these models can generate images of the same content, such as a car, under a variety of different styles and environments, e.g.\ \emph{real}, \emph{cartoon}, and \emph{night-time}. Due to this disentangled representation, the segmentation head learns more domain-invariant relations between the underlying features and the scene semantics, such as `car'. When given an image from a different domain, the diffusion-based segmentation model (Fig.~\ref{fig:motivation}(b), right) is therefore able to more robustly identify and segmenting the object, compared to utilizing a standard backbone with a more entangled representation (Fig.~\ref{fig:motivation}(a)).
These observations motivate us to explore the use of diffusion representations for DG and TTDA.

One key feature that distinguishes diffusion models from other backbones for semantic segmentation is their unique ability to manipulate the backbone using \emph{prompt conditioning}. This unique feature grants us direct control over domains, enabling us to generalize and adapt to new domains directly with parameter-efficient prompts. In this work, we aim at designing \textbf{\emph{simple yet effective}} methods for boosting DG and TTDA performance, without resorting to intricate techniques such as image translation, augmentation, or rare class sampling~\cite{hoffman2018cycada,yang2020fda,zhu2017unpaired,hoyer2022daformer}. To this end, we explore how to utilize the prompt in order to achieve even better generalization, or to adapt to new domains.

\textbf{Domain Generalization:} To improve the domain generalization ability of diffusion pretraining semantic segmentation models, we introduce category prompts and scene prompts as conditioning inputs to distinguish domain-invariant features from domain-variant ones. In addition, we propose a prompt randomization strategy to further improve the extraction and disentanglement of domain-invariant representations. This strategy ensures prediction consistency on the same image under different scene prompts, thereby enhancing the robustness of the model to domain shifts.

\textbf{Test-Time Domain Adaptation:} In order to facilitate adaptation of diffusion pretraining semantic segmentation models to the target domain during test time, we propose utilizing the scene prompt as the modulation parameter, which can be optimized via a loss function based on pseudo-labels during inference. The prompt tuning opens a new avenue for TTDA, which is parameter-efficient and mitigates the risk of overfitting.

To summarize, our contributions are four-fold:
\begin{itemize}[leftmargin=*]
	\setlength{\itemsep}{0pt}
	\setlength{\parsep}{-2pt}
	\setlength{\parskip}{-0pt}
	\setlength{\leftmargin}{-15pt}
	\vspace{-7pt}
        \item We conduct the first analysis of the generalization performance of diffusion pretrained models for semantic segmentation, demonstrating its superior performance.
        \item We introduce prompt-based methods, namely \emph{scene prompt} and \emph{prompt randomization}, to further improve the model's domain generalization capability.
        \item We propose a prompt tuning method to perform test-time domain adaptation of the model.
        \item Extensive experiments on four benchmarks demonstrate the effectiveness of our proposed approach. Notably, our DG and TTDA methods achieve 61.2\% and 62.0\% on Cityscapes $\rightarrow$ ACDC, even surpassing the SOTA unsupervised domain adaptation (UDA) DAFormer~\cite{hoyer2022daformer} by over 5.8 points.
\end{itemize}

\vspace{-5pt}\section{Related Work}
\textbf{Domain Generalization.} 
Previous approaches for domain generalization can be categorized into two main strategies: 1) image augmentation and 2) feature normalization and whitening. The first strategy involves randomly stylizing or augmenting images from the source domain, a technique known as domain randomization~\cite{yue2019domain,tremblay2018training,tobin2017domain,fan2022normalization}, to learn domain-invariant representations. The second strategy focuses on normalizing and whitening the features~\cite{li2017universal,ulyanov2017improved,choi2021robustnet,pan2018two,peng2022semantic} to ensure robustness across different domains. In contrast to these previous methods, our approach differs by not relying on stylized or translated images or perturbed features. Instead, we solely regulate the behavior of the model backbone through the use of prompts. This new approach allows us to achieve domain generalization without the need for extensive image transformations or feature manipulations.

\textbf{Test-Time Domain Adaptation.} 
Previous TTDA methods, also known as source-free domain adaptation~\cite{wang2020tent,mirza2022norm}, often focus on tuning the parameters of batch normalization layers, which are parameter-efficient. However, this approach has limitations in terms of adaptability and compatibility with network architectures other than convolutional neural networks, such as transformers~\cite{wang2022continual}. Alternatively, some methods optimize the entire model or its main components, such as the feature backbone~\cite{liu2021source,ye2021source}. However, such approaches tend to be parameter-heavy, making them prone to catastrophic overfitting to the noisy unsupervised learning objective, especially when the quantity of target domain data is limited. In contrast, our prompt-based method not only offers greater parameter efficiency compared to tuning batch normalization layers, but it also effectively modulates the behavior of the model's backbone.

\vspace{-5pt}\section{Method}
\subsection{Preliminaries}\label{sec:pre}
\textbf{Problem Statement.} \emph{Test-time domain adaptation (TTDA):} The objective of TTDA is to adapt a model $f_{\theta^s}$, with pre-trained with parameters $\theta^s$, on a labeled source domain dataset $\{\x^s, \y^s\}$ in order to improve the performance on the unlabeled target domain $\{\x^t\}$. The adaptation $\theta^s\rightarrow\theta^t$ is performed post-training, without access to the source domain data. \emph{Domain generalization (DG):} DG aims to generalize the model $f_{\theta^s}$, trained on the labeled source domain data $\{\x^s, \y^s\}$, to the unseen target domain $\{\x^t\}$, but without updating the model parameters $\theta^s$.

\textbf{Diffusion Models.} 
Diffusion models learn the reverse diffusion process to effectively convert a pure noise into a sample of the learned distribution. 
The forward diffusion process~\cite{ho2020denoising} gradually adds Gaussian noise to the input data $\z_0$ until it follows a simple Gaussian prior distribution, 
\begin{align}
    \label{eq:forwardnoise}
    \z_p &= \sqrt{\bar{\alpha}_p}\z_{0} + \sqrt{1-\bar{\alpha}_p}\epsilon\,,\quad \epsilon\sim\cN(\mathbf{0}, \I)\,,\quad \bar{\alpha}_p = \prod_{q=0}^{p}\alpha_{q} .
\end{align}
Here, $\z_p$ represents the latent feature variable at the $p$-th timestep and $\{\alpha_p\}$ are fixed coefficients that dictate the noise schedule.
Then, diffusion models employ a network $\epsilon_\theta$ that reverses the forward process by training it to estimate the noise $\epsilon$ which has been added to $\z_p$ in Eq.~\eqref{eq:forwardnoise}. This is done by minimizing a loss of the form,
\begin{eqnarray}
    \mathbb{E}_{p\sim\cU[1, P]}||\epsilon-\epsilon_\theta(\z_p, p; \cC)||^2 \,,
\end{eqnarray}
where $\cC$ represents an additional conditioning input to the network. In~\cite{zhao2023unleashing}, the conditioning input $\cC$ consists of $M$ tokens, derived from a text or an image prompt. 

\vspace{-5pt}\subsection{Diffusion-Pretraining for Semantic Segmentation} \label{sec:diff_seg}
A diffusion model trained for large-scale high-resolution image synthesis needs to learn both low-level (\eg object color and texture) and high-level knowledge (\eg object interactions and scene layout). Furthermore, text-guided diffusion models need to capture and relate the semantics conveyed in both the prompt and generated image. These observations led to the recent exploration of frozen diffusion models as the underlying pre-trained representation for semantic segmentation tasks~\cite{zhao2023unleashing}. The aforementioned work aims to extract and transfer the semantic relevant knowledge from the large-scale text-to-image synthesis pretraining to the downstream semantic segmentation task. The basic idea is to 1) utilize the pretrained diffusion model as the backbone network, 2) extract the visual internal representations $\{\f_i(\epsilon_\theta, \x^s)\}$, and cross-attention maps $\{\a_i(\f_i, \cC)\}$ between the conditioning input $\cC$ and the internal visual representations, and 3) feed the extracted $\{\f_i(\epsilon_\theta, \x^s)\}$ and $\{\a_i(\f_i, \cC)\}$ into a learned 
semantic projection head $\cD$, to obtain the predicted semantic segmentation map $\hat{\y}^s$,
\begin{equation}
    \hat{\y}^s = \cD(\f_i(\epsilon_\theta, \x^s), \a_i(\f_i, \cC)) \label{eq:pred_seg}
\end{equation}
Then, the semantic projection head $\cD$ is trained with the standard cross entropy loss, $\cL_{s} = CE(\hat{\y}^s, \y^s)$. During the training, the diffusion model $\epsilon_\theta$ is frozen and only the semantic projection head $\cD$ is optimized, \ie $\min_{\cD} \cL_s$.

\vspace{-5pt}\subsection{Prompting Diffusion Representations for Domain Generalization} \label{sec:dg}
\begin{figure*}
    \centering
    \includegraphics[width=\linewidth]{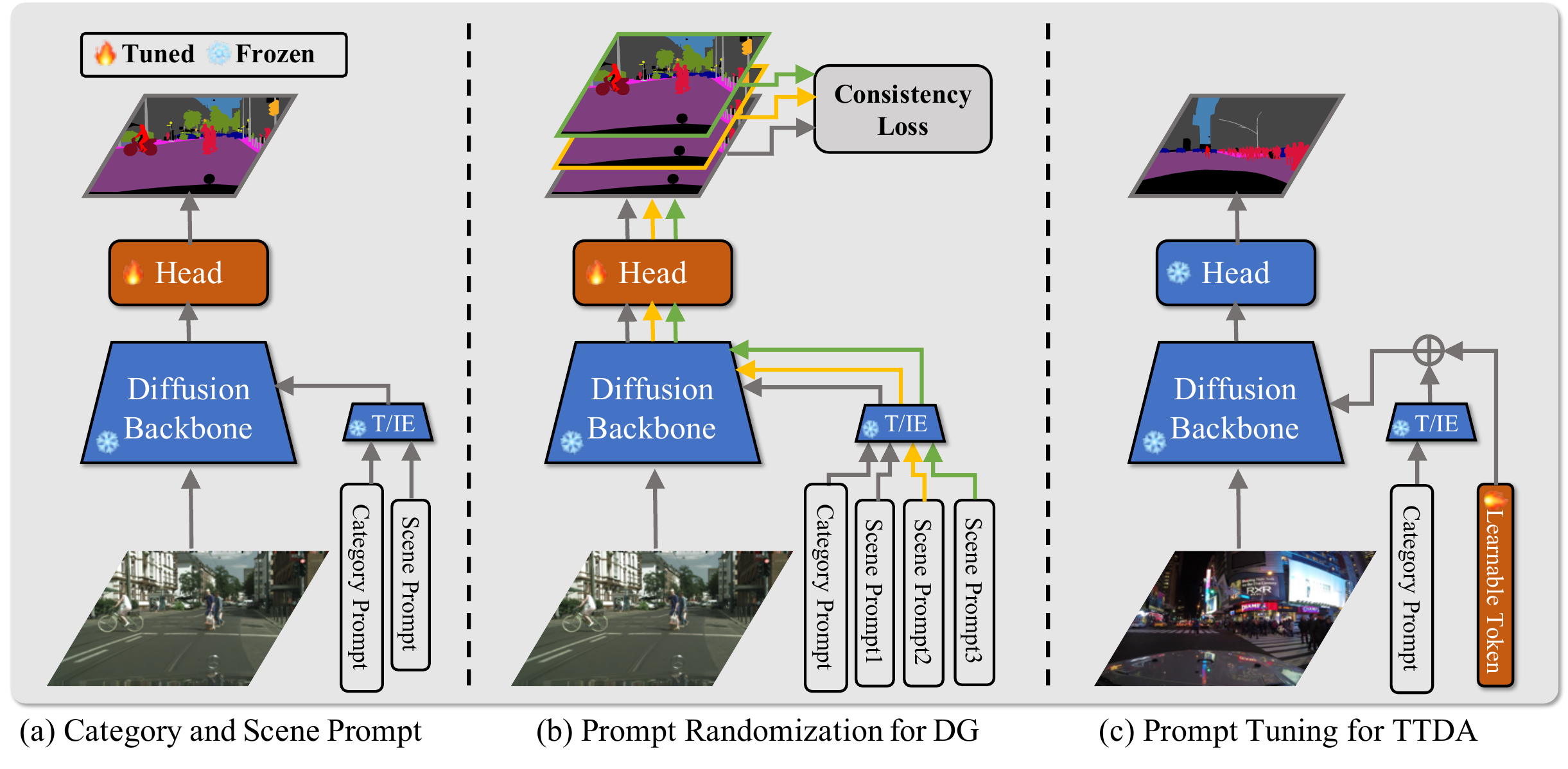}
    \vspace{-20pt}
    \caption{\textbf{Method Overview.} (a) We employ a frozen diffusion-pretrained backbone and a trained segmentation head. Our backbone is conditioned on a \emph{category prompt} and the introduced the \emph{scene prompt}, which are first input to the text/image prompt encoder ``T/IE". The prompts aids the extraction of domain-invariant representations (Sec.~\ref{sec:dg_cat_scene}). (b) We improve domain generalization by sampling random scene prompts and enforcing consistent predictions during training, facilitating learning of domain-invariant relations (Sec.~\ref{sec:dg_rand}). (c) For TTDA, we further propose a \emph{prompt tuning} strategy, which adapts the representation by learning the scene prompt token (Sec.~\ref{sec:ttda_pt}).}
    \vspace{-10pt}
    \label{fig:method_overview}
\end{figure*}

\subsubsection{Generalization Capabilities of Diffusion-Pretraining} \label{sec:dg_pretrain}
Given the remarkable success of diffusion-pretraining for semantic segmentation, a natural inquiry arises: \emph{To what extent do diffusion-pretrained segmentation models maintain their effectiveness under severe domain shift?} Motivated by this question, we first set out to investigate the diffusion-pretrained segmentation model's generalization performance in face of significant domain shift. In Table~\ref{tab:pretrain}, we compare popular pretraining strategies: (1) supervised pretraining for image classification on ImageNet-22k~\cite{deng2009imagenet}; (2) self-supervised pretraining for pixel reconstruction on ImageNet-1k~\cite{deng2009imagenet}; (3) CLIP pretraining, consisting of contrastive visual-language pairing~\cite{radford2021learning}; and (4) diffusion-driven pretraining for text-to-image synthesis on LAION-5B~\cite{schuhmann2022laion}. We analyze their generalization performance on GTA$\rightarrow$Cityscapes by comparing it to the \emph{oracle} performance for each network. The oracle is the same network trained in a fully supervised manner on the target Cityscapes dataset.
Note that the reported oracle mIoU values may appear lower than those in the literature on supervised learning~\cite{liu2021swin,xie2021segformer,liu2022convnet,he2022masked,dosovitskiy2020image}, as we follow the established convention of prior DG and TTDA works that downsample the Cityscapes images by a factor of two, to ensure fair comparison. In all cases, we assess the model's performance on the Cityscapes validation set by reporting the mean intersection-over-union (mIoU). To evaluate the generalization ability of each model, we present the relative mIoU compared to the oracle, following \cite{hoyer2022daformer}. Interestingly, higher oracle performance does not necessarily equate to better generalization on unseen target domains. This indicates that certain backbone models struggle with overfitting and do not effectively capture domain-invariant knowledge.

\begin{table}[t]
    \centering
    \caption{\textbf{Comparison to other pretraining methods}, under GTA $\rightarrow$ Cityscapes.}
    \resizebox{\textwidth}{!}{%
    \begin{tabular}{l|ccccc|c}
    \toprule
    \rowcolor[gray]{0.85}
    Architecture & Swin-B\cite{liu2021swin} & MiT-B5\cite{xie2021segformer} & ConvNeXt-B\cite{liu2022convnet} & MAE-ViT-L/16\cite{he2022masked} & CLIP-ViT-B\cite{radford2021learning} & Stable-Diffusion\cite{rombach2022high}\\
    \midrule
    Pretraining Type & \emph{supervised} & \emph{supervised} & \emph{supervised} & \emph{self-supervised} & \emph{visual-language} & \emph{text-to-image}\\
    Pretraining Dataset & \emph{ImageNet-22k} & \emph{ImageNet-22k} & \emph{ImageNet-22k} & \emph{ImageNet-1k} & \emph{CLIP} & \emph{LAION-5B}\\ \midrule
    Generalization & 38.9 & 45.6 & 46.0 & 42.7 & 39.0 & \textbf{49.2}\\
    Oracle & 79.2 & 76.4 & 79.9 & 76.8 & 70.0 & 74.7\\
    Relative & 49.1\% & 59.7\% & 57.6\% & 55.6\% & 55.7\% & \textbf{65.9\%}\\
    \bottomrule
    \end{tabular}%
    }
    \vspace{-15pt}
    \label{tab:pretrain}
\end{table}

However, we observe that the model using diffusion pretraining achieves superior performance in both absolute (49.2 mIoU) and relative (65.9\% of the oracle) generalization metrics. This demonstrates an exceptional generalization ability compared to other pretraining strategies. This remarkable generalization performance reached by the vanilla diffusion pretrained segmentation model, encourages us to further investigate their potential benefits in domain adaptation and generalization. 

The purpose of this work is to develop \emph{simple yet effective} method for domain adaptation and generalization problems, without any complex tricks, such as image translation, data augmentation and class sampling. Building upon the characteristics of diffusion models discussed in Sec.~\ref{sec:pre} and Sec.~\ref{sec:diff_seg}, we note that these models are distinguished by their capacity to be finely controlled by the conditioning input $\cC$, derived from image or text prompts. Different from previous methods, that change the backbone behaviors by modulating specific networks layers, stylizing images or introducing additional networks, prompts tuning opens a new avenue of manipulating the backbones representation in a effective and efficient way. In the next sections, we propose novel prompt tuning methods for domain generalization and test-time domain adaptation, based on diffusion-pretrained semantic segmentation models. An overview of our prompt-based approach is depicted in Fig.~\ref{fig:method_overview}.

\vspace{-5pt}\subsubsection{Category and Scene Prompt} \label{sec:dg_cat_scene}
To improve the generalization ability of diffusion-pretrained segmentation models, we first introduce the \emph{category prompt} $\cC_c$ and the \emph{scene prompt} $\cC_s$ as the conditioning inputs $\cC = [\cC_c; \cC_s]$. These are used to disentangle the domain-invariant features, such as object classes and scene layout, and the domain-variant features, such as object color and texture, scene style and lighting. To further harvest and utilize the domain-invariant knowledge to enhance the generalization ability, we propose the prompt randomization strategy during training, to enforce consistency of predictions on the same image and category prompt amidst varying scene prompts. 

\textbf{Category Prompt.} The category prompt is typically defined as a template of "\texttt{a photo of a [Class]}", where "\texttt{[Class]}" are \emph{category names} (\eg road, sidewalk and sky for the street scene image). \textit{I.e., } the category prompt only provides the class names as the guidance, to extract the domain-invariant knowledge. For instance, using the "car" class as an example, diffusion models can synthesize car images with varying attributes by providing different prompts. However, despite the diverse attribute inputs, the fundamental identity of the object as a car remains unchanged as long the prompts include "\texttt{a photo of a car}". This highlights the ability of category prompts to effectively capture knowledge related to the object's core identity, i.e.\ ``what is a car?'', from other attributes, such as color or body type. The core identity of the object is domain-invariant and exactly what the domain generalization needs.
For the $C-$class semantic segmentation, the category prompts are $C$ tokens, each of which is $N-$dim vector.

\textbf{Scene Prompt.} 
The category prompt can capture the main features of an object that stay the same across different scenarios, \ie domain-invariant knowledge. To better extract domain invariant representations, we further condition the network on an introduced scene prompt. Our hypothesis is that the diffusion network can better extract domain-invariant representations if it is aware of the image domain. Consider e.g.\ a night photo of a street scene. It might be difficult to recognize objects, such as cars, pedestrians, and buildings in such conditions. However, by making the diffusion representation explicitly aware of the conditions through a style prompt ``A dark night photo'', we believe that it can partly revoke the domain-specific effect as it will explicitly consider a night-time view of a car, pedestrian, or building. Thus, to further facilitate the extraction of domain-invariant knowledge across different domains, we introduce the scene prompt, $\cC_s$.
One example of scene prompt is a template "\texttt{a [scene] photo}", \eg "a GTA5 photo" or "a night photo".

By combining the category prompt $\cC_c$ and the scene prompt $\cC_s$ as the conditional inputs, the predicted semantic segmentation map in Eq.~(\ref{eq:pred_seg}) is rewritten as,
\begin{eqnarray}
    \hat{\y}^s = \cD(\f_i(\epsilon_\theta, \x^s), \a_i(\f_i, [\cC_c;\cC_s])) \label{eq:pred_seg_full}
\end{eqnarray}
Note that the scene prompt $\cC_s$ can not only be defined as the aforementioned text template, but also be designated as a $N-$dim learnable prompt, or an image prompt obtained by feeding an example image into pretrained language-image encoder (\eg CLIP~\cite{radford2021learning}). With the category and scene prompts employed, there are $M=C+1$ tokens of $N-$dim vector in total as the conditioning inputs $\cC$.

\vspace{-5pt}\subsubsection{Prompt Randomization for Domain Generalization} \label{sec:dg_rand}
By incorporating the category and scene prompts as conditional inputs, the diffusion-pretraining segmentation model is able to extract domain-invariant knowledge, leading to enhanced generalization ability. To further capture the domain-invariant knowledge and boost the domain generalization capabilities, we propose a prompt randomization strategy. Our idea is to enforce consistency between the semantic predictions under different scene prompts $\{\cC_s^k\}_{k=1}^{K}$. The intuition is that a model capable of generalizing well would make similar predictions for images containing the same content, irrespective of their domain-variant attributes, such as weather or style. 

By feeding various scene prompts $\{\cC_s^k\}_{k=1}^{K}$ into the diffusion backbone in Sec.~\ref{sec:dg_cat_scene}, the corresponding semantic segmentation maps are obtained as $\{\hat{\y}^s_k\}_{k=1}^{K}$, where $\hat{\y}^s_k = \cD(\f_i(\epsilon_\theta, \x^s), \a_i(\f_i, [\cC;\cE_k]))$. Then, the consistency loss, $\cL_c$, between different scene prompts are formulated as,
\begin{eqnarray}
\cL_c = \sum_{p,q\in\{1,...,K\},q\neq p} KL(\hat{\y}^s_p||\hat{\y}^s_q) = -\sum_{p,q\in\{1,...,K\},q\neq p} \hat{\y}^s_p\log\frac{\hat{\y}^s_p}{\hat{\y}^s_q},
\end{eqnarray}
where $KL(\cdot||\cdot)$ represents the Kullback–Leibler (KL) divergence~\cite{cover1999elements}, which aligns the semantic prediction under different scene prompts. The complete learning objective for prompt randomization is the combination of the semantic segmentation loss $\cL_s$ and the consistency loss $\cL_c$, written as,
\begin{eqnarray}
    \cL_{total} = \sum_{k=1}^{K}CE(\hat{\y}^s_k, \y^s) + \lambda\cL_c.
\end{eqnarray}
Here, $\lambda$ is the hyper-parameter used to balance the semantic segmentation loss and the consistency loss, which is set to 0.1 in this work.

\vspace{-5pt}\subsection{Prompting Diffusion Representations for Test-Time Domain Adaptation}\label{sec:ttda}
\subsubsection{Test-Time Domain Adaptation} \label{sec:ttda_pre} 
In Sec.~\ref{sec:dg}, we introduce category and scene prompts to extract domain-invariant knowledge and improve the generalization ability of the diffusion-pretraining semantic segmentation model across different domains. The domain generalization strategy utilizes only the labeled source domain data ${\x^s, \y^s}$ without accessing the unlabeled target domain data $\x^t$. A natural next question is thus: \emph{Can the model be effectively adapted given the unlabeled target domain data during test-time, \ie test-time domain adaptation?} Following that, we examine how well the diffusion-pretraining semantic segmentation models can adapt to new domains at test-time, by leveraging $\x^t$.

Test-time domain adaptation (TTDA) presents two main challenges that must be addressed: (1) how can the source-domain initialized model be \textbf{modulated} effectively in light of unsupervised learning objective fraught with noise? (2) What \textbf{learning objectives} should be adopted to enable optimization if only unlabeled data from the target domain is provided? Our work primarily addresses challenge (1), and employs a pseudo-label based optimization objective for challenge (2) as it is proven simple yet effective in the test-time domain adaptation field. Next, we propose the prompt tuning based method for test-time domain adaptation, with diffusion pretraining semantic segmentation models.

\subsubsection{Prompt Tuning for Test-Time Domain Adaptation} \label{sec:ttda_pt}
\textbf{Modulation Parameters.} 
To effectively tackle the aforementioned challenge (1) in TTDA, the main focus is on identifying the relevant parameters that need to be updated in order to control the behavior of the backbone in a desirable manner. Our diffusion pretraining models, described in Sec.~\ref{sec:dg}, leverage the category and scene prompts to effectively control the behavior of the backbone. 
More specifically, the \emph{category prompt}, $\cC_c$, captures \emph{domain-invariant} knowledge on the object core identity, shared by the source and target domains. The \emph{scene prompt}, $\cC_s$, introduces the domain-specific information to further help disentangeling the representation. Test-time domain adaptation involves a domain shift from the source to the target domain. Therefore, the scene prompts needs to be updated to accommodate this shift in domains. The basic idea of our prompt tuning for TTDA is to learn the scene prompt $\cC_s$ to facilitate adaptation from the source to the target domain. That is, the scene prompt serves as the modulation parameter, which can be updated by $\theta^t\leftarrow \theta^s :\cE\leftarrow \cC_s+\Delta\cC_s$.

\textbf{Learning Objective.} Our test-time optimization objective $\cL_t$ is to tune the scene prompt $\cE$ supervised by the pseudo-label $\tilde{\y}^t=\argmax \cD(\f_i(\epsilon_\theta, \x^t), \a_i(\f_i, [\cC_c;\cC_s]))$, formulated as,
\begin{equation}
    \cL_t = CE(\hat{\y}^t, \tilde{\y}^t)\,, \qquad \cC_s\leftarrow \cC_s+\partial \cL_t/\partial \cC_s.
\end{equation}
The only optimized parameters during test-time is the scene prompt $\cC_s$, which is a $N-$dim vector and set as 768-dim in this work. Thus, our prompt-tuning method for TTDA is parameter-efficient, enabling fast adaptation and helping to mitigate the risk of overfitting during the TTDA process.

\vspace{-5pt}\section{Experiments} \label{sec:exp}
\subsection{Experimental Setup and Implementation Details}
\textbf{Datasets.} We evaluate the effectiveness of our proposed prompt-based method for DG and TTDA under different scenarios, including synthetic-to-real and clear-to-adverse benchmarks. We use the conventional notation A$\rightarrow$B to describe the DG and TTDA task, where A and B are source and target domain, respectively. \emph{Synthetic-to-Real}: There are two settings, GTA~\cite{richter2016playing} $\rightarrow$ Cityscapes~\cite{cordts2016cityscapes} and SYNTHIA~\cite{ros2016synthia} $\rightarrow$ Cityscapes~\cite{cordts2016cityscapes}. \emph{Clear-to-Adverse}: There are also two tasks, Cityscapes~\cite{cordts2016cityscapes} $\rightarrow$ ACDC~\cite{sakaridis2021acdc} and Cityscapes~\cite{cordts2016cityscapes} $\rightarrow$ Dark Zurich~\cite{sakaridis2019guided}. For ease of reference, we use the following abbreviations throughout the text: G$\rightarrow$C, S$\rightarrow$C, C$\rightarrow$A, and C$\rightarrow$D, respectively. More detailed description about different datasets is put in the supplementary.

\textbf{Implementation Details.} \emph{Backbone and Semantic Segmentation Head}: We use the released v1-5 version of Stable Diffusion~\cite{rombach2022high} as the backbone, and the Semantic FPN~\cite{kirillov2019panoptic} decoder as the segmentation head. \emph{Prompts}: We leverage the publicly available pretrained ViT-L/14 CLIP~\cite{radford2021learning} model to map text/image prompts into the feature space that can be utilized by Stable Diffusion. \emph{Scene Prompt}: The scene prompt for prompt randomization by default is composed of two components: (1) the text description for the source domain, such as "a GTA5 photo" for the GTA dataset, and (2) the text description for the target domain, such as "a night photo" for the Dark Zurich dataset, called the text prompt version. As an alternative ot (2), we also experiment with an image from the target domain, \ie the image prompt version. Specific prompts used for each experiment are put in the supplementary. \emph{Test-Time Domain Adaptation}: The model is firstly pre-trained on the source domain with both category prompt and scene prompt (see Sec.~\ref{sec:dg_cat_scene} and Source($\cC_s$) in Table~\ref{tab:ablation}), and then updated during the test-time with our proposed prompt tuning strategy in Sec.~\ref{sec:ttda_pt}.

\vspace{-5pt}\subsection{Comparison with state-of-the-art}

\textbf{Domain Generalization.} In Sec.~\ref{sec:dg_rand}, we propose the prompt randomization strategy for DG with diffusion pretraining models. As shown in Table~\ref{tab:SOTA_DG}, our prompt randomization method is demonstrated to outperform previous SOTA DG methods by a significant margin. Notably, scene prompts for prompt randomization can be obtained flexibly in different types, including text (DG-T) and image (DG-I) prompts. And both types are proven effective, improving the vanilla diffusion models (Van.) performance significantly.
\begin{table*}[t]
    \caption{\textbf{Comparison to SOTA DG methods.} $^\dagger$ denotes results obtained from~\cite{schwonberg2023augmentation}.}
    \vspace{-5pt}
    \begin{subtable}{0.49\linewidth}
    \centering
    \caption{Synthetic-to-Real.}
    \vspace{-5pt}
    \resizebox{\linewidth}{!}{%
    \begin{tabular}{l|cccc}
    \rowcolor[gray]{0.85}
    \toprule
    Method & Backbone & Extra Data & G$\rightarrow$C & S$\rightarrow$C\\
    \midrule
    \midrule
    IBN-Net\cite{pan2018two} & ResNet-101 & \xmark & 37.4 & 34.2\\
    DRPC\cite{yue2019domain} & ResNet-101 & \checkmark & 42.5 & 37.6\\
    ISW\cite{choi2021robustnet} & ResNet-101 & \xmark & 37.2 & 37.2$^\dagger$ \\
    FSDR\cite{huang2021fsdr} & ResNet-101 & \checkmark & 44.8 & 40.8 \\
    GTR\cite{peng2021global} & ResNet-101 & \checkmark & 43.7 & 39.7\\
    SAN-SAW\cite{peng2022semantic} & ResNet-101 & \xmark & 45.3 & 40.9  \\
    SHADE\cite{zhao2022style} & ResNet-101 & \xmark & 46.7 & - \\
    WEDGE\cite{kim2021wedge} & ResNet-101 & \checkmark & 45.2 & 40.9 \\
    \midrule
    AugFormer-S\cite{schwonberg2023augmentation} & MiT-B5 & \xmark & 45.6 & 40.3\\
    AugFormer\cite{schwonberg2023augmentation} & MiT-B5 & \xmark & - & 44.2$^\dagger$\\
    \midrule
    \midrule
    Ours (Van.) & \emph{Diffusion} & \xmark & 49.2 & 47.8\\
    Ours (DG-T) & \emph{Diffusion} & \xmark & \textbf{52.0} & 49.1 \\
    Ours (DG-I) & \emph{Diffusion} & \xmark & \textbf{52.0} & \textbf{49.3} \\
    \bottomrule
    \end{tabular}
    }
    \end{subtable}
    \begin{subtable}{0.5\linewidth}
    \centering
    \caption{Clear-to-Adverse (val set).}
    \vspace{-5pt}
    \resizebox{\linewidth}{!}{%
    \begin{tabular}{l|cccc}
    \rowcolor[gray]{0.85}
    \toprule
    Method & Backbone & Extra Data & C$\rightarrow$A & C$\rightarrow$D\\
    \midrule
    \midrule
    ISA\cite{li2023intra} & ResNet-101 & \xmark & 47.4 & 26.1\\
    ISW+ISA\cite{li2023intra} & ResNet-50 & \xmark & 47.6 & 23.1\\
    MixS~\cite{zhou2021domain} & ResNet-101 & \xmark & 37.0 & 9.4\\ 
    MixS+ISA\cite{li2023intra} & ResNet-101 & \xmark & 41.8 & 20.6 \\
    DSU\cite{li2022uncertainty} & ResNet-101 & \xmark & 38.3 & 12.3 \\
    DSU+ISA\cite{li2023intra} & ResNet-101 & \xmark & 43.3 & 24.6\\
    IBN-Net~\cite{pan2018two} & ResNet-50 & \xmark & 44.1 & 21.7 \\
    ISW+MSA\cite{reddy2022master} & ResNet-50 & \xmark & 47.3 & 22.5 \\
    ISW+MSA~\cite{reddy2022master} & ResNet-101 & \xmark & 49.0 & 24.8 \\
    SiamDoGe~\cite{wu2022siamdoge} & ResNet-50 & \xmark & 52.3 & - \\
    
    \midrule
    \midrule
    Ours (Van.) & \emph{Diffusion} & \xmark & 57.0 & 31.2\\
    Ours (DG-T) & \emph{Diffusion} & \xmark & \textbf{58.6} & \textbf{34.0} \\
    Ours (DG-I) & \emph{Diffusion} & \xmark & 58.4 & \textbf{34.0}\\
    \bottomrule
    \end{tabular}
    }
    \end{subtable}
    \label{tab:SOTA_DG}%
    \vspace{-15pt}
\end{table*}

\textbf{Test-time domain adaptation.}
In Sec.~\ref{sec:ttda_pt}, we propose the prompt tuning strategy to adapt the diffusion representation during test time. The results in Table~\ref{tab:ttda} demonstrate the superior performance of our prompt tuning method for TTDA, achieving a remarkable improvement of 5.9\%, 6.5\%, 2.7\%, and 4.9\% over existing TTDA methods on different benchmarks.
\begin{table*}[b]
    \vspace{-10pt}
    \caption{\textbf{Comparison to SOTA TTDA methods.} ``Param. Eff." represents parameter efficient. $^\dagger$~represents the baseline is combined with~\cite{wang2020tent} and higher resolution image is used for training.}
    \vspace{-5pt}
    \begin{subtable}{0.49\linewidth}
    \centering
    \caption{Synthetic-to-Real.}
    \vspace{-5pt}
    \resizebox{\linewidth}{!}{%
    \begin{tabular}{l|cccc}
    \rowcolor[gray]{0.85}
    \toprule
    Method & Backbone & Param. Eff. & G$\rightarrow$C & S$\rightarrow$C\\
    \midrule
    \midrule
    TransAdapt\cite{das2023transadapt} & ResNet-101 & \checkmark & 37.8 & 33.7\\
    Tent\cite{wang2020tent} & ResNet-101 & \checkmark & 38.9 & 35.5 \\
    SFDA\cite{liu2021source} & ResNet-50 & \xmark & 43.2 & 39.2 \\
    SFUDA\cite{ye2021source} & ResNet-101 & \xmark & 49.4 & 44.2 \\
    URMA\cite{fleuret2021uncertainty} & ResNet-101 & \xmark & 45.1 & 39.6\\
    SHOT\cite{liang2020we} & ResNet-101 & \xmark & 44.1 & -\\
    AUGCO\cite{prabhu2021augco} & ResNet-50 & \checkmark & 47.1 & 39.5 \\
    HCL\cite{huang2021model} & ResNet-101 & \xmark & 48.1 & 43.5\\
    C-SFDA\cite{karim2023c} & ResNet-101 & \xmark & 48.3 & 44.6 \\
    CO-SFDA\cite{karim2023c} & ResNet-101 & \checkmark & 46.3 & 43.0 \\
    \midrule
    \midrule
    Ours (Van.) & \emph{Diffusion} & -- & 49.2 & 47.8\\
    Ours (TTDA) & \emph{Diffusion} & \checkmark & \textbf{52.2} & \textbf{49.5} \\
    \bottomrule
    \end{tabular}
    }
    \end{subtable}
    \begin{subtable}{0.492\linewidth}
    \centering
    \caption{Clear-to-Adverse.}
    \vspace{-5pt}
    \resizebox{\linewidth}{!}{%
    \begin{tabular}{l|c|cccc}
    \rowcolor[gray]{0.85}
    \toprule
    Method & Set & Backbone & Param. Eff. & C$\rightarrow$A & C$\rightarrow$D\\
    \midrule
    \midrule
    TTBN\cite{nado2020evaluating} & \multirow{7}{*}{\rotatebox{90}{Val-Set}}& ResNet-101 & \checkmark & - & 28.0\\
    TENT\cite{wang2020tent} & &ResNet-101 & \checkmark & - & 26.6\\
    AUGCO\cite{prabhu2021augco} & &ResNet-101 & \checkmark & - & 32.4\\
    CO-SFDA\cite{karim2023c} & &ResNet-101 & \checkmark & - & 33.2\\
    MSA\cite{reddy2022master} & & ResNet-101 & \checkmark & 47.9 & 22.8 \\
    MSA\cite{reddy2022master} & & MiT-B0 & \checkmark & 46.6 & 20.2\\
    \cmidrule{1-1}\cmidrule{3-6}
    Ours (Van.) & & \emph{Diffusion} & -- & 57.0 & 31.2\\
    Ours (TTDA) & & \emph{Diffusion}  & \checkmark & \textbf{58.5} & \textbf{37.0} \\
    \midrule
    \midrule
    TENT & \multirow{5}{*}{\rotatebox{90}{Test-Set}} & ResNet-101 & \checkmark & 49.0 & - \\
    HCL & & ResNet-101 & \checkmark & 46.8 & - \\
    URMA & & ResNet-101 & \checkmark & 47.2 & - \\
    SegFormer~\cite{xie2021segformer} $^\dagger$ & & MiT-B5 & \xmark & 59.3 & 42.8\\
    \cmidrule{1-1}\cmidrule{3-6}
    Ours (TTDA) & & \emph{Diffusion} & \checkmark & \textbf{62.0} & \textbf{47.7} \\
    \bottomrule
    \end{tabular}
    }
    \end{subtable}
    \label{tab:ttda}%
\end{table*}

\textbf{Unsupervised domain adaptation.} 
As show in Table~\ref{tab:uda}, our proposed DG methods (DG-T and DG-I) and TTDA method exhibit exceptional performance, outperforming even the strong unsupervised domain adaptation (UDA) methods that have access to both the source and target domain at the same time for training. Notably, our DG method achieves a performance gain of 5.8\% over the strong UDA method, DAFormer, on the Cityscapes$\rightarrow$ACDC benchmark, despite not utilizing any data from ACDC. Fig.~\ref{fig:seg_vis} presents qualitative comparisons among our DG, TTDA, and DAFormer results, illustrating the effectiveness of our prompt-based method. For instance, when examining the "road" class, we observe that without prompt conditioning, DAFormer segments the "road" into the upper sky region. In contrast, our category prompt conditioning, specifically "a photo of a road," assists the model in learning that the "road" should appear in the lower part of the image rather than the upper sky region, thus preventing this error from occurring.
\begin{table*}[t]
    \centering
    \caption{\textbf{Comparison between our DG, TTDA method to UDA methods,} under C$\rightarrow$A. All methods are evaluated on the test set through the online public evaluation server. $^\dagger$ represents the use of extra auxiliary reference images that are geographically aligned and captured under clear-weather/daytime.}
    \resizebox{\textwidth}{!}{%
    \begin{tabular}{l|l|ccccccccccccccccccc|c}
    \rowcolor[gray]{0.85}
    \toprule
    Setting&Method & Road&SW&Build&Wall&Fence&Pole&TL&TS&Veg&Terrain&Sky&Person&Rider&Car&Truck&Bus&Train&MC&Bike& mIoU\\
    \midrule
    \rowcolor[gray]{0.95}
    \multicolumn{21}{c}{Cityscapes $\rightarrow$ ACDC (\textbf{Test Set})} \\
    \midrule
    \multirow{5}{*}{\textbf{UDA}}&ADVENT\cite{vu2018advent} & 72.9 & 14.3 & 40.5 & 16.6 & 21.2 & 9.3 & 17.4 & 21.2 & 63.8 & 23.8 & 18.3 & 32.6 & 19.5 & 69.5 & 36.2 & 34.5 & 46.2 & 26.9 & 36.1 & 32.7 \\
    &GCMA\cite{sakaridis2019guided} $\dagger$ & 79.7 & 48.7 & 71.5 & 21.6 & 29.9 & 42.5 & 56.7 & 57.7 & 75.8 & 39.5 & 87.2 & 57.4 & 29.7 & 80.6 & 44.9 & 46.2 & 62.0 & 37.2 & 46.5 & 53.4 \\
    &MGCDA\cite{sakaridis2020map} $\dagger$ & 73.4 & 28.7 & 69.9 & 19.3 & 26.3 & 36.8 & 53.0 & 53.3 & 75.4 & 32.0 & 84.6 & 51.0 & 26.1 & 77.6 & 43.2 & 45.9 & 53.9 & 32.7 & 41.5 & 48.7 \\
    &DANNet\cite{wu2021dannet} $\dagger$ & 84.3 & 54.2 & 77.6 & 38.0 & 30.0 & 18.9 & 41.6 & 35.2 & 71.3 & 39.4 & 86.6 & 48.7 & 29.2 & 76.2 & 41.6 & 43.0 & 58.6 & 32.6 & 43.9 & 50.0 \\
    &DAFormer\cite{hoyer2022daformer} & 58.4 & 51.3 & 84.0 & 42.7 & 35.1 & 50.7 & 30.0 & 57.0 & 74.8 & 52.8 & 51.3 & 58.3 & 32.6 & 82.7 & 58.3 & 54.9 & 82.4 & 44.1 & 50.7 & 55.4 \\
    \midrule
    \multirow{2}{*}{\textbf{DG}}&Ours (DG-T) & 89.6 & 62.5 & 84.4 & 48.6 & 39.9 & 49.2 & 48.7 & 55.6 & 74.5 & 48.3 & 86.1 & 60.3 & 39.9 & 84.9 & 62.6 & 63.5 & 73.6 & 37.7 & 52.3 & \textbf{61.2}\\
    &Ours (DG-I) & 91.1 & 85.9 & 84.7 & 83.8 & 76.4 & 66.3 & 66.1 & 62.6 & 56.4 & 56.1 & 54.6 & 54.4 & 51.4 & 50.6 & 49.6 & 47.8 & 42.2 & 41.1 & 38.8 & 61.0\\ \midrule
    \textbf{TTDA}&Ours & 88.2 & 86.0 & 85.6 & 85.5 & 74.0 & 73.8 & 62.3 & 61.2 & 60.0 & 57.6 & 56.5 & 55.9 & 52.3 & 52.0 & 50.8 & 50.3 & 42.4 & 42.2 & 41.8 & \textbf{62.0}\\
    \bottomrule
    \end{tabular}
    }
    \label{tab:uda}%
    \vspace{-5pt}
\end{table*}

\vspace{-5pt}\subsection{Ablation Study and Prompts Analysis} \label{sec:exp_ablation}
\textbf{Different Scene Prompts Comparison.} We conduct ablation experiments to evaluate the effectiveness of our proposed scene prompt in improving the domain generalization performance of diffusion pretraining models. \emph{With and Without Scene Prompt}: First, we compare the performance of models with and without scene prompt to verify its effectiveness. As shown in Table~\ref{tab:ablation}, the models with any of the different scene prompts (target, learned, and source) all outperformed the baseline model without scene prompt, achieving mIoU scores of 50.9\%, 51.4\%, and 51.4\% respectively, compared to 49.2\% for the baseline on the GTA$\rightarrow$Cityscapes benchmark. \emph{Different Scene Prompts}: Next, we investigate the impact of different scene prompt choices by comparing the prompts obtained from 1) text description of target domain, referred to as "Target"; 2) text description of source domain, referred to as "Source"; and 3) a learnable parameter, referred to as "Learned". Our results show that the "Source" scene prompt outperforms the "Target" and "Learned" prompts on different benchmarks. This finding confirms our statement in Sec.~\ref{sec:dg_cat_scene} that the scene prompt is used to disentangle domain-invariant knowledge and revoke the effect of domain-variant factors in the source domain. Hence, the "Source" prompt, which captures domain-variant factors in the source domain, works best.

\textbf{Increased Number of Class Prompts.} To assess the impact of classes in category prompts, we conduct an experiment in which more classes were utilized in the category prompt. More specifically, in all GTA$\rightarrow$Cityscapes experiments in this work, the category prompt consisting of 19 classes is used. To obtain the category prompt with more classes, we utilize 150 classes from the ADE20K dataset~\cite{zhou2017scene}. This expanded set of classes not only includes the 19 object classes used in our standard category prompt, but also encompasses a variety of additional classes. Our analysis in Table~\ref{tab:pt_analysis} demonstrates that increasing the number of classes in the category prompt leads to a improvement in the generalization ability of diffusion pretraining semantic segmentation models, 50.1\% \vs 49.2\%. This improvement can be attributed to the fact that providing a greater number of classes as the category prompt enables the diffusion representations to more accurately distinguish between a broader range of class objects, thereby mitigating the issue of mis-classification. This finding suggests a promising direction for future work to further improve the generalization ability of diffusion pretraining semantic segmentation models by incorporating more auxiliary classes into the category prompt.
\begin{minipage}{\textwidth}
  \begin{minipage}[b]{0.43\textwidth}
    \centering
    \vspace{5pt}
    \includegraphics[width=\linewidth]{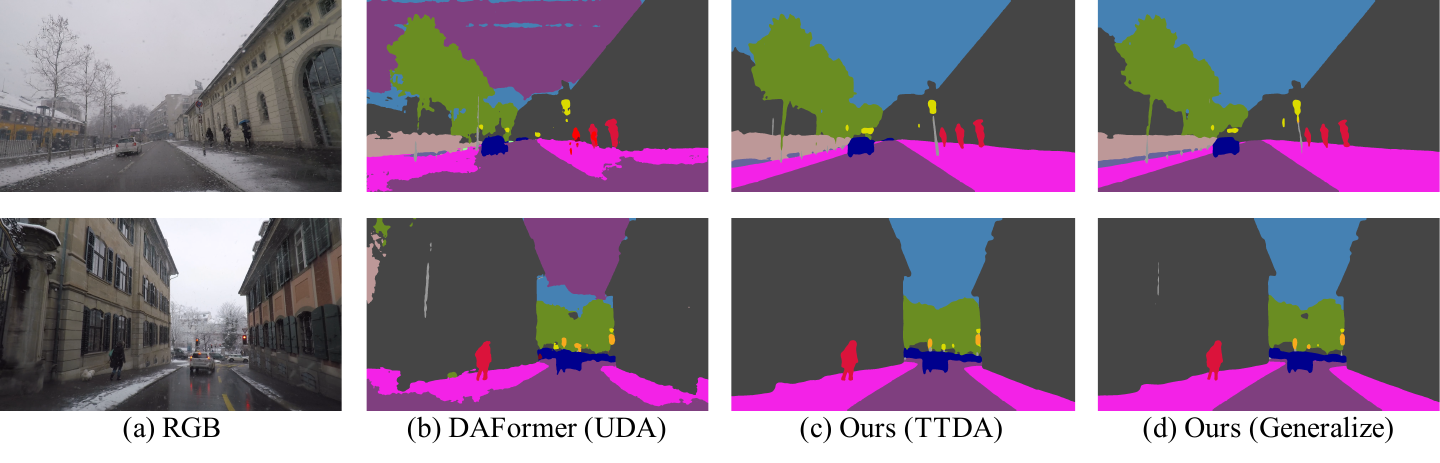}
    \vspace*{-18pt}
    \captionof{figure}{Qualitative comparisons between our DG, TTDA and DAFormer~\cite{hoyer2022daformer} method. DAFormer is a UDA method, which uses the full target domain data during training.}
    \label{fig:seg_vis}
  \end{minipage}
  \hfill
  \begin{minipage}[b]{0.55\textwidth}
    \centering
    \captionof{table}{\textbf{Ablation study and comparisons between different prompts types,} under synthetic-to-real and clear-to-adverse benchmarks. Results are evaluated on the val-set of target domain.}
    \vspace{-5pt}
    \setlength{\tabcolsep}{1.0pt}
    \resizebox{\linewidth}{!}{%
    \begin{tabular}[b]{l|c|ccc|ccc}
    \rowcolor[gray]{0.85}
    \toprule
    Setting & w/o $\cC_s$ & Target ($\cC_s$) & Learned ($\cC_s$) & Source ($\cC_s$) & TTDA & DG-T & DG-I\\
    \midrule
    G$\rightarrow$C & 49.2 & 50.9 & 51.4 & 51.4 & 52.2 & 52.0 & 52.0 \\
    S$\rightarrow$C & 47.8 & 48.4 & 48.2 & 48.8 & 49.5 & 49.1 & 49.3\\
    C$\rightarrow$D & 31.2 & 32.2 & 30.4 & 32.8 & 37.0 & 34.0 & 34.0\\
    C$\rightarrow$A & 57.0 & 57.0 & 58.0 & 58.0 & 58.5 & 58.6 & 58.4\\
    \bottomrule
    \end{tabular}
    \label{tab:ablation}
    }
    \end{minipage}
  \end{minipage}
\begin{table*}[t]
    \centering
    \caption{\textbf{Prompts analysis on,} 1) increased number of category prompts (150 classes); 2) scene prompts (a ``water, grass, sand, painting" photo) irrelevant to source/target domain, under G $\rightarrow$ C.}
    \resizebox{\textwidth}{!}{%
    \begin{tabular}{l|ccccccccccccccccccc|c}
    \rowcolor[gray]{0.85}
    \toprule
    Method & Road&SW&Build&Wall&Fence&Pole&TL&TS&Veg&Terrain&Sky&Person&Rider&Car&Truck&Bus&Train&MC&Bike& mIoU\\
    \midrule
    \rowcolor[gray]{0.95}
    \multicolumn{21}{c}{\emph{\text{Increased number of \textbf{category prompts}}}} \\
    \midrule
    19\text{ Classes} & 70.9 & 25.7 & 88.1 & 54.5 & 43.6 & 43.5 & 46.3 & 32.7 & 87.8 & 52.1 & 90.9 & 63.0 & 24.9 & 87.7 & 34.5 & 42.6 & 2.9 & 22.2 & 20.8 &49.2\\
    \midrule
    150\text{ Classes} & 76.0 & 29.9 & 88.2 & 41.5 & 41.3 & 45.0 & 45.2 & 29.5 & 87.8 & 52.4 & 90.2 & 63.7 & 27.7 & 86.9 & 34.6 & 47.3 & 8.1 & 28.6 & 28.4 & 50.1\\
    \midrule
    \midrule
    \rowcolor[gray]{0.95}
    \multicolumn{21}{c}{\emph{\text{Prompt randomization with irrelevant \textbf{scene prompts}}}} \\
    \midrule
    Irrelevant& 85.6 & 36.7 & 87.8 & 52.7 & 44.9 & 41.7 & 45.8 & 31.4 & 87.5 & 51.2 & 89.7 & 64.1 & 29.5 & 87.4 & 29.7 & 35.7 & 13.0 & 31.7 & 36.9 & 51.7 \\
    DG-Text & 87.7 & 36.2 & 87.7 & 43.3 & 38.2 & 38.1 & 44.4 & 31.2 & 87.6 & 48.3 & 89.8 & 63.9 & 31.7 & 89.4 & 61.9 & 50.6 & 0.1 & 24.5 & 33.9 & 52.0\\
    \bottomrule
    \end{tabular}
    }
    \vspace{-10pt}
    \label{tab:pt_analysis}%
\end{table*}

\textbf{Prompt Randomization with Irrelevant Prompts.} In our primary experiments, the scene prompt is generated from text/image prompts that are relevant to the target domain. However, to evaluate the flexibility and scalability of the scene prompt, we conduct an experiment in which the scene prompt is generated from a random, unrelated scene description. For instance, in the GTA$\rightarrow$Cityscapes experiment, we used scene prompts such as "a sand photo," "a grass photo," "a painting photo," and "a water photo." The results presented in Table~\ref{tab:pt_analysis} demonstrate that the prompt randomization method using irrelevant prompts performs favorably in comparison to the method using target-relevant prompts, achieving 51.7\% and 52.0\%, respectively.

\vspace{-5pt}\section{Conclusion}
In this work, we conduct the first study on the generalization performance of a semantic segmentation model utilizing pretrained diffusion representations, demonstrating their superior performance compared to other pretraining backbones. To further enhance the model's domain generalization capability, we introduce novel prompt-based methods: the scene prompt and prompt randomization. Additionally, we propose a prompt tuning method that enables efficient and effective test-time domain adaptation of the model. Through extensive experiments conducted on four benchmarks, we validate the effectiveness of our proposed simple yet powerful approach.

\textbf{Limitations.}
The current approach employs hand-designed prompts.
An interesting future direction is to leverage other large language models to automatically generate accurate prompts for our method.

{\small
\bibliographystyle{ieee_fullname}
\bibliography{egbib}
}



\section*{Supplementary material}

# Prompting Diffusion Representations for Cross-Domain Semantic Segmentation
# —Supplementary Material—

Rui Gong[1]    Martin Danelljan[1]    Han Sun[2]    Julio Delgado Mangas[3]    Luc Van Gool[1]
[1]CVL, ETH Zürich  [2] IMOS, EPFL  [3] Meta Reality Labs
{gongr,martin.danelljan,vangool}@vision.ee.ethz.ch
han.sun@epfl.ch  julio.delgadomangas@gmail.com

In this supplementary material, we provide the additional information for,

- **S1** discussion on the potential societal impact,
- **S2** implementation and training details of our proposed method,
- **S3** detailed information about datasets involved in our experiments,
- **S4** additional experimental results on more ablation study, combination with rare class sampling technique, and extension to depth estimation.

## S1   Societal Impact

Our proposed approach has the potential to adapt the semantic segmentation model to unseen unlabeled data, offering significant cost and effort savings by eliminating the need for manual labeling when dealing with new data and requirements. However, this advantage also brings a potential downside of reduced demand for data labeling jobs, which could result in unemployment within this domain.

## S2   Method Implementation and Training Details

In the main paper, we introduce the diffusion representations prompting method for addressing cross-domain semantic segmentation problems. In this supplementary material, we provide comprehensive implementation and training details of our proposed method, offering a more in-depth understanding of its practical aspects.

**Scene Prompts.** In Table S1, the specific category and scene prompt for each experiment is listed.

**Training Details.** *Batch Size:* In all experiments presented in our paper, we use a batch size of 2. *Optimizer and Parameters:* We employ the AdamW optimizer [10] with a learning rate of 8e-5 and a weight decay of 5e-3. The learning rate warming-up iterations are set to 1500, and the total training iterations are set to 80000. We utilize a polynomial learning rate scheduler with a power of 0.9 and a minimum learning rate of 1e-5. *Compute Resources:* The code implementation is based on PyTorch [11]. All experiments are conducted on an NVIDIA Tesla V100 GPU with 32GB memory. The training time for prompt randomization experiments is approximately 32 hours, while for other domain generalization experiments, it takes around 16 hours to complete the total 80000 training iterations.

## S3   Datsets Information

As introduced in Sec. 4 of the main paper, our experiments involve five datasets: GTA [13], SYNTHIA [16], Cityscapes [5], Dark Zurich [17] and ACDC [18]. In Sec. S4.4, our experiments on depth

Table S1: **Category and scene prompt list for each experiment.** For category prompt, "C classes" represents C tokens, each of which is a template of `"a photo of a [Class]"` and `"[Class]"` are *category names*.

| Experiment | Table Number | Benchmark | Category Prompt | Scene Prompt | mIoU |
|---|---|---|---|---|---|
| Ours (Van.) | Table 2, 3 | G→C | 19 classes | – | 49.2 |
| Ours (Van.) | Table 2, 3 | S→C | 19 classes | – | 47.8 |
| Ours (DG-T) | Table 2, 5 | G→C | 19 classes | "a GTA photo", "a European street scene photo" | 52.0 |
| Ours (DG-T) | Table 2, 5 | S→C | 19 classes | "an unreal rendered photo", "a European street scene photo" | 49.3 |
| Ours (DG-I) | Table 2, 5 | G→C | 19 classes | "a GTA photo", *prompt encoded from a Cityscapes image* | 52.0 |
| Ours (DG-I) | Table 2, 5 | S→C | 19 classes | "an unreal rendered photo", *prompt encoded from a Cityscapes image* | 49.1 |
| Ours (Van.) | Table 2, 3 | C→A | 19 classes | – | 57.0 |
| Ours (Van.) | Table 2, 3 | C→D | 19 classes | – | 31.2 |
| Ours (DG-T) | Table 2, 4, 5 | C→A | 19 classes | "a European street scene photo", "a night photo", "a fog photo", "a rain photo", "a snow photo" | 58.6 |
| Ours (DG-T) | Table 2, 5 | C→D | 19 classes | "a European street scene photo", "a night photo" | 34.0 |
| Ours (DG-I) | Table 2, 4, 5 | C→A | 19 classes | "a European street scene photo", *prompt encoded from a night, fog, rain and snow image, resp.* | 58.4 |
| Ours (DG-I) | Table 2, 5 | C→D | 19 classes | "a European street scene photo", *prompt encoded from a night, fog, rain and snow image, resp.* | 34.0 |
| 19 classes | Table 6 | G→C | 19 classes | – | 49.2 |
| 150 classes | Table 6 | G→C | 150 classes | – | 50.1 |
| Irrelevant | Table 6 | G→C | 19 classes | "a GTA photo", "a sand photo", "a water photo", "a grass photo", "a painting photo" | 51.7 |
| DG-Text | Table 6 | G→C | 19 classes | "a GTA photo", "a European street scene photo" | 52.0 |
| w/o $\mathcal{C}_s$ | Table 5 | G→C | 19 classes | – | 49.2 |
| w/o $\mathcal{C}_s$ | Table 5 | S→C | 19 classes | – | 47.8 |
| w/o $\mathcal{C}_s$ | Table 5 | C→A | 19 classes | – | 57.0 |
| w/o $\mathcal{C}_s$ | Table 5 | C→D | 19 classes | – | 31.2 |
| Target($\mathcal{C}_s$) | Table 5 | G→C | 19 classes | "a European street scene photo" | 50.9 |
| Target($\mathcal{C}_s$) | Table 5 | S→C | 19 classes | "a European street scene photo" | 48.4 |
| Target($\mathcal{C}_s$) | Table 5 | C→A | 19 classes | "a night, fog, rain or snow photo" | 57.0 |
| Target($\mathcal{C}_s$) | Table 5 | C→D | 19 classes | "a night photo" | 32.2 |
| Learned($\mathcal{C}_s$) | Table 5 | G→C | 19 classes | *learned token* | 51.4 |
| Learned($\mathcal{C}_s$) | Table 5 | S→C | 19 classes | *learned token* | 48.2 |
| Learned($\mathcal{C}_s$) | Table 5 | C→A | 19 classes | *learned token* | 58.0 |
| Learned($\mathcal{C}_s$) | Table 5 | C→D | 19 classes | *learned token* | 30.4 |
| Source($\mathcal{C}_s$) | Table 5 | G→C | 19 classes | "a GTA photo" | 51.4 |
| Source($\mathcal{C}_s$) | Table 5 | S→C | 19 classes | "an unreal rendered photo" | 48.8 |
| Source($\mathcal{C}_s$) | Table 5 | C→A | 19 classes | "a European street scene photo" | 58.0 |
| Source($\mathcal{C}_s$) | Table 5 | C→D | 19 classes | "a European street scene photo" | 32.8 |

estimation involve two datasets, NYU-Depth v2 [19] and HyperSim [14]. In the following sections, we provide detailed information about each dataset.

**GTA.** The GTA [13] dataset consists of synthetic urban-scene images generated using a game engine. It contains 24,966 images, each with a resolution of 1914×1052 pixels, and pixel-wise semantic segmentation annotations. The urban scenes in the GTA dataset are designed to resemble the layout of Los Angeles, representing a typical U.S. urban environment. To align with previous domain generalization and test-time domain adaptation works [21, 12], we resize the GTA images to 1280×720.

**SYNTHIA.** SYNTHIA [16] is a synthetic photo-realistic image dataset generated from a virtual city environment. In our experiments, we utilize the SYNTHIA-RAND-Cityscapes dataset, which is specifically designed for street scene parsing tasks and contains 9,400 densely labeled images. These images have a resolution of 1280×760 pixels, which we keep for our experiments.

**Cityscapes.** Cityscapes [5] is a dataset comprising real street-scene images captured from diverse European cities. The training set of Cityscapes consists of 2,975 images, while the validation set contains 500 images. The images in Cityscapes have a resolution of 2048×1024 pixels. In line with previous works on domain generalization and test-time domain adaptation [21, 12], we resize the Cityscapes images to 1024×512 pixels to ensure consistency with the experimental setup.

**Dark Zurich.** Dark Zurich [17] is a dataset comprising real nighttime urban-scene images captured in Zurich. The training, validation and test set of Dark Zurich include 2,416, 50 and 151 images. It is important to note that access to the ground truth for the test set of Dark Zurich is limited to the online benchmark, and it is not publicly available. The images in Dark Zurich have a resolution of 1920×1080 pixels.

**ACDC.** ACDC [18] is a real-world street-scene image dataset specifically designed to capture adverse weather conditions such as fog, snow, rain, as well as nighttime scenes. It consists of a total of 4,006 images, including 1,600 training images, 406 validation images, and 2,000 test images, all with a resolution of 1920×1080 pixels. Similar to the evaluation process for Dark Zurich, the evaluation of the ACDC dataset's test set can only be performed using the online benchmark. The ground truth for the test set is not publicly available.

**NYU-Depth v2.** NYU-Depth v2 [19] is a real-world indoor-scene dataset which is captured by both RGB and depth cameras from the Microsoft Kinect. The dataset contains 24K training samples, with depth values ranging from 1e-3m to 10m.

Table S2: **Ablation study and comparisons between different prompts types,** under synthetic-to-real and clear-to-adverse benchmarks.

| Method | Road | SW | Build | Wall | Fence | Pole | TL | TS | Veg | Terrain | Sky | Person | Rider | Car | Truck | Bus | Train | MC | Bike | mIoU |
|---|---|---|---|---|---|---|---|---|---|---|---|---|---|---|---|---|---|---|---|---|
| | | | | | | | | | GTA → Cityscapes | | | | | | | | | | | |
| w/o $\mathcal{C}_s$ | 70.9 | 25.7 | 88.1 | 54.5 | 43.6 | 43.5 | 46.3 | 32.7 | 87.8 | 52.1 | 90.9 | 63.0 | 24.9 | 87.7 | 34.5 | 42.6 | 2.9 | 22.2 | 20.8 | 49.2 |
| Target ($\mathcal{C}_s$) | 78.3 | 30.1 | 87.8 | 46.0 | 41.4 | 40.4 | 44.6 | 32.7 | 87.6 | 49.6 | 89.8 | 62.1 | 18.8 | 88.2 | 53.7 | 49.3 | 1.3 | 36.6 | 28.3 | 50.9 |
| Learned ($\mathcal{C}_s$) | 77.9 | 28.4 | 88.2 | 47.3 | 43.3 | 44.2 | 46.8 | 32.4 | 88.1 | 53.9 | 90.3 | 64.0 | 27.1 | 87.7 | 35.8 | 53.2 | 2.5 | 34.1 | 31.6 | 51.4 |
| Source ($\mathcal{C}_s$) | 73.0 | 28.9 | 87.8 | 45.8 | 40.8 | 41.3 | 46.0 | 30.5 | 87.6 | 49.2 | 89.2 | 62.8 | 21.5 | 88.3 | 52.6 | 63.7 | 5.3 | 31.7 | 29.9 | 51.4 |
| TTA | 86.1 | 36.2 | 87.8 | 47.0 | 39.9 | 42.0 | 46.1 | 36.5 | 87.9 | 49.5 | 90.1 | 64.6 | 28.4 | 88.2 | 39.4 | 54.2 | 3.3 | 31.3 | 32.7 | 52.2 |
| DG-Text | 87.7 | 36.2 | 87.7 | 43.3 | 38.2 | 38.1 | 44.4 | 31.2 | 87.6 | 48.3 | 89.8 | 63.9 | 31.7 | 89.4 | 61.9 | 50.6 | 0.1 | 33.9 | 33.9 | 52.0 |
| DG-Image | 82.8 | 32.6 | 88.4 | 52.3 | 43.8 | 44.1 | 46.2 | 32.5 | 87.8 | 51.6 | 90.8 | 64.3 | 31.3 | 88.5 | 35.5 | 50.8 | 2.7 | 32.0 | 30.1 | 52.0 |
| | | | | | | | | | SYNTHIA → Cityscapes | | | | | | | | | | | |
| w/o $\mathcal{C}_s$ | 77.1 | 34.8 | 85.4 | 19.5 | 2.2 | 46.6 | 25.8 | 35.1 | 83.9 | - | 87.8 | 62.3 | 27.7 | 83.4 | - | 40.7 | - | 23.0 | 29.0 | 47.8 |
| Target ($\mathcal{C}_s$) | 78.0 | 34.9 | 85.6 | 22.6 | 2.7 | 46.4 | 27.8 | 43.3 | 82.2 | - | 85.3 | 60.5 | 30.5 | 85.5 | - | 34.3 | - | 23.6 | 30.4 | 48.4 |
| Learned ($\mathcal{C}_s$) | 85.9 | 44.9 | 85.9 | 13.3 | 3.7 | 47.1 | 21.4 | 40.0 | 82.8 | - | 85.9 | 61.3 | 28.6 | 79.6 | - | 32.6 | - | 26.8 | 30.9 | 48.2 |
| Source ($\mathcal{C}_s$) | 81.2 | 38.5 | 85.3 | 21.0 | 3.9 | 48.0 | 22.2 | 35.4 | 83.4 | - | 87.1 | 60.5 | 29.6 | 85.5 | - | 43.5 | - | 27.1 | 28.4 | 48.8 |
| TTA | 87.2 | 46.6 | 86.0 | 20.5 | 3.5 | 47.7 | 18.3 | 34.0 | 85.0 | - | 90.0 | 61.7 | 28.9 | 85.7 | - | 44.3 | - | 26.3 | 26.5 | 49.5 |
| DG-Text | 89.9 | 52.2 | 85.6 | 20.1 | 3.4 | 47.2 | 25.7 | 34.1 | 83.3 | - | 87.0 | 59.8 | 27.9 | 82.5 | - | 40.0 | - | 16.9 | 29.7 | 49.1 |
| DG-Image | 82.1 | 39.1 | 85.7 | 24.4 | 3.3 | 46.3 | 27.1 | 36.4 | 84.3 | - | 88.6 | 61.5 | 32.0 | 84.0 | - | 38.2 | - | 27.0 | 28.0 | 49.3 |
| | | | | | | | | | Cityscapes → Dark Zurich | | | | | | | | | | | |
| w/o $\mathcal{C}_s$ | 84.1 | 44.3 | 74.9 | 43.6 | 44.4 | 36.8 | 9.6 | 21.9 | 43.0 | 13.2 | 19.9 | 22.7 | 31.5 | 61.8 | 0.0 | 0.0 | 0.0 | 5.1 | 35.1 | 31.2 |
| Target ($\mathcal{C}_s$) | 87.6 | 53.5 | 66.0 | 46.8 | 54.7 | 36.6 | 22.2 | 26.4 | 40.1 | 16.9 | 8.9 | 25.2 | 34.3 | 61.4 | 0.0 | 0.0 | 0.0 | 14.2 | 16.4 | 32.2 |
| Learned ($\mathcal{C}_s$) | 86.7 | 53.6 | 76.9 | 34.1 | 28.8 | 41.1 | 1.1 | 18.0 | 42.3 | 17.6 | 20.0 | 20.9 | 42.2 | 58.5 | 0.0 | 0.0 | 0.0 | 14.1 | 21.7 | 30.4 |
| Source ($\mathcal{C}_s$) | 85.3 | 45.8 | 73.8 | 46.6 | 42.3 | 40.5 | 6.6 | 25.3 | 43.2 | 24.9 | 17.5 | 23.4 | 41.7 | 60.7 | 0.0 | 0.0 | 0.0 | 12.1 | 34.4 | 32.8 |
| TTA | 84.6 | 45.2 | 78.7 | 47.2 | 42.7 | 42.7 | 4.1 | 26.4 | 60.2 | 24.2 | 75.1 | 23.8 | 41.1 | 59.1 | 0.0 | 0.0 | 0.0 | 12.7 | 35.2 | 37.0 |
| DG-Text | 86.9 | 53.3 | 82.3 | 50.2 | 46.0 | 36.4 | 13.9 | 22.3 | 37.7 | 28.5 | 5.8 | 21.3 | 29.3 | 64.1 | 0.0 | 0.0 | 0.0 | 33.8 | 33.8 | 34.0 |
| DG-Image | 84.0 | 40.3 | 73.7 | 38.6 | 43.2 | 42.0 | 7.2 | 22.8 | 52.7 | 34.1 | 46.4 | 23.3 | 43.0 | 64.7 | 0.0 | 0.0 | 0.0 | 12.1 | 17.8 | 34.0 |
| | | | | | | | | | Cityscapes → ACDC | | | | | | | | | | | |
| w/o $\mathcal{C}_s$ | 87.8 | 54.9 | 78.4 | 50.7 | 42.7 | 53.8 | 52.7 | 53.2 | 74.9 | 35.8 | 84.0 | 52.3 | 24.2 | 82.5 | 73.9 | 58.1 | 56.1 | 30.5 | 36.1 | 57.0 |
| Target ($\mathcal{C}_s$) | 86.8 | 52.7 | 77.0 | 47.9 | 41.7 | 55.3 | 58.1 | 50.4 | 75.2 | 36.8 | 82.5 | 50.6 | 25.6 | 83.1 | 69.4 | 58.2 | 59.0 | 33.0 | 39.7 | 57.0 |
| Learned ($\mathcal{C}_s$) | 89.5 | 61.3 | 81.7 | 47.2 | 39.1 | 56.1 | 61.8 | 52.1 | 72.5 | 35.9 | 82.2 | 51.2 | 28.6 | 82.1 | 66.5 | 61.2 | 58.1 | 35.7 | 39.7 | 58.0 |
| Source ($\mathcal{C}_s$) | 88.0 | 55.5 | 79.1 | 48.6 | 41.4 | 56.3 | 57.0 | 53.0 | 73.6 | 38.0 | 81.5 | 54.6 | 33.3 | 82.0 | 71.7 | 60.8 | 54.9 | 35.1 | 38.7 | 58.0 |
| TTA | 88.1 | 56.1 | 81.2 | 49.9 | 41.0 | 56.3 | 58.5 | 53.8 | 71.8 | 38.5 | 81.6 | 55.7 | 33.4 | 82.9 | 72.5 | 61.7 | 55.6 | 34.7 | 37.4 | 58.5 |
| DG-Text | 89.6 | 61.7 | 79.7 | 45.4 | 38.7 | 54.8 | 58.8 | 54.1 | 73.9 | 37.6 | 83.4 | 50.7 | 30.2 | 82.9 | 71.5 | 62.6 | 60.1 | 32.6 | 45.5 | 58.6 |
| DG-Image | 91.7 | 66.1 | 81.3 | 48.7 | 40.7 | 55.5 | 55.7 | 52.3 | 73.2 | 37.0 | 82.0 | 43.7 | 29.4 | 81.9 | 79.9 | 63.9 | 63.9 | 27.0 | 36.2 | 58.4 |

Table S3: **Ablation study on category prompt,** under GTA → Cityscapes benchmark.

| Method | Road | SW | Build | Wall | Fence | Pole | TL | TS | Veg | Terrain | Sky | Person | Rider | Car | Truck | Bus | Train | MC | Bike | mIoU |
|---|---|---|---|---|---|---|---|---|---|---|---|---|---|---|---|---|---|---|---|---|
| w/o $\mathcal{C}_c$ | 80.9 | 31.3 | 87.2 | 37.4 | 39.2 | 41.3 | 44.6 | 28.1 | 87.7 | 52.1 | 89.8 | 63.5 | 19.5 | 89.0 | 38.2 | 44.4 | 2.3 | 22.1 | 21.1 | 48.4 |
| w/ $\mathcal{C}_c$ | 70.9 | 25.7 | 88.1 | 54.5 | 43.6 | 43.5 | 46.3 | 32.7 | 87.8 | 52.1 | 90.9 | 63.0 | 24.9 | 87.7 | 34.5 | 42.6 | 2.9 | 22.2 | 20.8 | 49.2 |

**HyperSim.** HyperSim [14] is a photo-realistic synthetic dataset for holistic scene understanding. The dataset contains 7,690 samples for evaluation, with depth values ranging from from 1e-3m to 80m.

# S4 Experiments

## S4.1 More detailed Ablation Study on Scene Prompt

In Section 4.3 of the main paper, we conducted the ablation study on the scene prompt and compared the performance of different scene prompts. Here, in Table S2, we present the detailed per-class IoU results corresponding to the ablation study and comparisons. It is shown that the scene prompt utilizing the text description of the source domain works the best, and the models with scene prompt all outperform that without scene prompt.

## S4.2 Additional Ablation Study on Category Prompt

In Sec. 4.3 and Sec. S4.1, we analyze the impact of the scene prompt and assess its contribution to the performance of domain generalization (DG) and test-time domain adaptation (TTDA). Additionally, we investigate the effect of the category prompt by conducting experiments on the GTA→Cityscapes benchmark, comparing models with and without the category prompt. TableS3 demonstrates that the model with the category prompt achieves better performance than the model without it, with mIoU scores of 49.2% and 48.4%, respectively. This improvement can be attributed to the category prompt's assistance in learning domain-invariant knowledge for challenging classes such as "wall", "fence", "pole", "traffic sign", and "rider".

Table S4: **Combination with rare class sampling (RCS),** under synthetic-to-real benchmarks.

| Method | Road | SW | Build | Wall | Fence | Pole | TL | TS | Veg | Terrain | Sky | Person | Rider | Car | Truck | Bus | Train | MC | Bike | mIoU |
|---|---|---|---|---|---|---|---|---|---|---|---|---|---|---|---|---|---|---|---|---|
| GTA → Cityscapes ||||||||||||||||||||||
| $\mathcal{C}_s + \mathcal{C}_c$ | 73.0 | 28.9 | 87.8 | 45.8 | 40.8 | 41.3 | 46.0 | 30.5 | 87.6 | 49.2 | 89.2 | 62.8 | 21.5 | 88.3 | 52.6 | 63.7 | 5.3 | 31.7 | 29.9 | 51.4 |
| $\mathcal{C}_s + \mathcal{C}_c$ + RCS | 89.0 | 45.8 | 88.2 | 53.6 | 43.3 | 45.2 | 46.7 | 32.1 | 88.0 | 49.9 | 90.0 | 65.3 | 36.8 | 88.1 | 36.8 | 53.4 | 7.3 | 29.4 | 37.4 | **54.0** |
| SYNTHIA → Cityscapes ||||||||||||||||||||||
| $\mathcal{C}_s + \mathcal{C}_c$ | 81.2 | 38.5 | 85.3 | 21.0 | 3.9 | 48.0 | 22.2 | 35.4 | 83.4 | - | 87.1 | 60.5 | 29.6 | 85.5 | - | 43.5 | - | 27.1 | 28.4 | 48.8 |
| $\mathcal{C}_s + \mathcal{C}_c$ + RCS | 88.7 | 49.8 | 85.1 | 19.8 | 4.9 | 47.8 | 25.0 | 42.5 | 82.7 | - | 85.9 | 59.7 | 28.6 | 83.3 | - | 37.1 | - | 25.3 | 36.9 | **50.2** |

Table S5: **Domain generalizable depth estimation performance,** under NYU-Depth v2 → HyperSim. "Diff. Rep." represents the vanilla diffusion representations based depth estimation model [23].

| Method | Backbone | RMSE↓ | $\log_{10}$ ↓ | REL↓ | $\delta_1$ ↑ |
|---|---|---|---|---|---|
| BTS [8] | *DenseNet* [7] | 6.404 | 0.329 | 0.476 | 0.225 |
| AdaBins [2] | *EfficientNet-B5* [20] | 6.546 | 0.345 | 0.483 | 0.221 |
| LocalBins [3] | *EfficientNet-B5* [20] | 6.362 | 0.320 | 0.468 | 0.234 |
| NewCRF [22] | *Swin-L* [9] | 6.017 | 0.283 | 0.442 | 0.255 |
| ZoeD-X-N [4] | *BEiT-L* [1] | 5.889 | 0.267 | **0.421** | **0.284** |
| Diff. Rep. | *Stable Diffusion* [15] | **5.145** | **0.262** | 0.448 | 0.268 |

### S4.3 Combination with Rare Class Sampling

The main paper demonstrates that our prompt-based diffusion representations significantly enhance domain generalization performance without the need for complex techniques like rare class sampling (RCS) [6]. In this section, we present additional experimental results that combine our approach with RCS, showcasing the flexibility of our method to be integrated with other techniques and further improve performance. TableS4 displays the domain generalization performance on the GTA→Cityscapes and SYNTHIA→Cityscapes benchmarks, where the integration of our category prompt, scene prompt, and RCS achieves performance of 54.0% and 50.2%, respectively. Rare class sampling is beneficial due to the GTA and SYNTHIA dataset attributes, specifically the presence of classes (*e.g.* bike, traffic sign, fence) that are relatively rare in GTA and SYNTHIA but more common in Cityscapes. This sampling strategy helps address the class imbalance issue and ensures that the model receives sufficient exposure to these rare classes during training, improving its ability to accurately recognize and segment them in the Cityscapes dataset.

### S4.4 Diffusion Representations for Generalizable Depth Estimation

In this paper, we showcase the robust domain generalization capabilities of diffusion representations for the cross-domain semantic segmentation task. This raises the question: *Can these domain generalization abilities be applied to other tasks as well?* While fully exploring this question requires further extensive research, we provide initial experimental results demonstrating the domain generalization performance of depth estimation. The performance of the vanilla diffusion representations-based depth estimation model is showcased in Table S5. It outperforms the previous methods, namely BTS [8], AdaBins [2], and LocalBins [3], across all metrics in the domain generalization scenario. When compared to the recent method NewCRF [22], the vanilla diffusion representations-based model achieves significantly better results in terms of RMSE, $\log_{10}$, and $\delta_1$, while achieving similar performance in terms of REL (0.448 vs 0.442). Moreover, it demonstrates competitive performance compared to the very recent SOTA method ZoeD-X-N [4], outperforming in RMSE and $\log_{10}$ metrics but slightly underperforming in REL and $\delta_1$ metrics. The notable achievement of the vanilla diffusion representations-based model is its RMSE score of 5.145, which establishes a new SOTA result and demonstrates significant improvement over the previous SOTA performance (5.145 vs 5.889, where lower values indicate better performance). The experimental findings provide compelling evidence that the efficacy of diffusion representations extends to diverse tasks, demonstrating their robust generalization capabilities.



\end{document}